\documentclass[a4paper,11pt]{amsart}
\usepackage{graphicx}
 \usepackage{mathptmx}
 \usepackage{amsaddr}
\usepackage{amsmath}
\usepackage{amsfonts}
\usepackage{amssymb}
\usepackage{enumitem}
\usepackage{multicol}
\usepackage{xcolor}
\usepackage{framed}
\usepackage{graphics}
\usepackage{kotex}
\usepackage{setspace}
\usepackage{lipsum} 
\usepackage{amsaddr} 
\usepackage[verbose=true]{microtype}
\usepackage[utf8]{inputenc}
\usepackage{amsmath}
\usepackage{lipsum}  
\usepackage{parskip} 
\usepackage{geometry} 
\usepackage{booktabs}
\usepackage{caption} 
\graphicspath{ {./images/} }

\newmuskip\pFqmuskip

\newcommand*\pFq[6][8]{%
  \begingroup 
  \pFqmuskip=#1mu\relax
  \mathcode`=\string"8000
  \begingroup\lccode`\~=`\,
  \lowercase{\endgroup\let~}\pFqcomma
  F^{#2}_{#3}{\left(\genfrac..{0pt}{}{#4}{#5}\bigg|#6\right)}%
  \endgroup
}
\newcommand{\pFqcomma}{\mskip\pFqmuskip}
\DeclareGraphicsExtensions{.pdf,.png,.jpg}

\begin{document}

\title[ ]{Statistical Analysis by Semiparametric Additive Regression and LSTM-FCN Based Hierarchical Classification for Computer Vision Quantification of Parkisonian Bradykinesia}

\author{Youngseo Cho$^{1,*}$, In Hee Kwak$^2$, Dohyeon Kim$^1$, Jinhee Na$^1$, Hanjoo Sung$^1$, Jeongjae Lee$^2$, Young Eun Kim$^{2,\dagger}$, Hyeo-il Ma$^{2,\dagger}$}

\address{$^1$VTOUCH Inc., Seoul, South Korea}
\address{$^2$Department of Neurology, Hallym University Sacred Heart Hospital, Hallym University College of Medicine, Anyang, South Korea}

\email{$^*$14choyoung@snu.ac.kr}
\email{$^\dagger$\{yekneurology@hallym.or.kr, hima@hallym.or.kr\}}

\subjclass[]{}
\maketitle

\section*{Abstract}
Bradykinesia, characterized by involuntary slowing or decrement of movement, is a fundamental symptom of Parkinson’s Disease (PD) and is vital for its clinical diagnosis. While various methodologies have been explored to quantify bradykinesia, computer vision-based approaches have shown promising results. However, these methods often fall short of adequately addressing key bradykinesia characteristics in the tepetitive limb movement: “occasional arrest” and “decrement in amplitude.” 
This research advances the vision-based quantification of bradykinesia by introducing a nuanced numerical analysis to capture decrement in amplitudes and employing a simple deep learning technique, LSTM-FCN for the precise classification of occasional arrests. Our approach structures the classification process hierarchically, tailoring it to the unique dynamics of bradykinesia in PD.
A statistical analysis of the extracted features, including those representing arrest and fatigue, has demonstrated their statistical significance in most cases. This finding underscores the importance of considering “occasional arrest” and “decrement in amplitude” in bradykinesia quantification of limb movement. Our enhanced diagnostic tool has been rigorously tested on an extensive dataset comprising 1396 motion videos from 310 PD patients, achieving an accuracy of 80.3\%. The results confirm the robustness and reliability of our method.

\bigskip
\bigskip

\section{INTRODUCTION}
Parkinson's Disease (PD)[1] is a neurodegenerative disorder primarily diagnosed and assessed based on movement disorders. The cardinal motor feature for the diagnosis of PD is bradykinesia, characterized by slowness of movement with a reduced or progressively diminishing amplitude or cessation of movement. The Movement Disorder Society-Unified Parkinson’s Disease Rating Scale (MDS-UPDRS)is the predominant tool employed for assessing the severity of clinical features, including bradykinesia in PD[2]. In the MDS-UPDRS part III, symptom severity is quantified on a scale from 0 to 4, with 0 indicating normality and 4 representing a complete inability to perform movements. Within the motor assessment of MDS-UPDRS part III, limb bradykinesia is evaluated through five specific tasks: Finger Tapping, Hand Movements, Rapid Alternating Movements, Toe Tapping, and Foot Stomping, with patients instructed to perform these tasks successively with the greatest possible speed and amplitude.
\begin{table*}[t]
 {\raggedright \textbf{Examiner Instructions:} Test each side separately.
Demonstrate the movement, but do not
continuously show it while the patient is being
tested. Instruct the patient to execute the motion as
quickly and as widely as possible 10 times.
Evaluate each side independently, assessing speed,
range of motion, arrest, stopping, and reduction in
movement amplitude.\par}~~~\\
\begin{tabular}{cl}
Score & Bradykinesia Criteria \\
\hline
0 & \begin{tabular}{l}Normal: No issues..\end{tabular} \\
\hline
1 & \begin{tabular}{l}
Slight: Any of the following: \\
a) Regular movement stops once or
twice or there is an arrest in the
movement;\\
 b) A bit of slowing; \\
c) Reduction in movement amplitude
towards the end of 10 actions. \\
\end{tabular} \\
\hline
2 & \begin{tabular}{l}
Mild: Any of the following:\\
a) 3-5 interruptions in the movement; \\
b)Mild slowing; \\
c) Reduction in movement amplitude around the middle of the 10 actions. \\
\end{tabular} \\
\hline
3 & \begin{tabular}{l}
Moderate: Any of the following: \\
a) More than 5 stops during the movement or at least one longer stop \\
\quad (freezing) during the movement; \\
b) Moderate slowing; \\
c) Reduction in movement amplitude starting after the first action.
\end{tabular} \\
\hline
4 & \begin{tabular}{l}
Severe: The movement is barely performable or not performable at all due to slowing,\\
 stopping, 
and reduced range of motion.
\end{tabular} \\
\hline
\end{tabular}\\
~~~\\
\caption{\emph{The criteria of limb bradykinesia in
MDS-UPDRS, amplitude and interval of movement,
arrest and fatigue are considered key factors.}}
\end{table*}
While the MDS-UPDRS provides a universal standard for assessment, the evaluation of PD motor features faces challenges due to the difficulty of accurately analyzing movements visually and the inherent subjectivity in clinicians' judgments[3]. Consequently, there has been a continuous effort to quantitatively overcome these limitations in PD diagnosis[4]-[5].

\smallskip

Among various approaches [4]-[13], computer vision-based quantification has shown outstanding results [5],[11]-[13]. Recent studies in this domain have leveraged advanced human pose estimation tools, like OpenPose[14], to generate time-series graphs describing patient motions. Features are then extracted from these graphs using techniques such as local extrema extraction, forming the basis for ordinal classification algorithms[15] in diagnostic machines.  
As human pose estimation and classification algorithms continue to advance, this methodology remains a promising area of research, particularly when considering the challenges of applying deep learning efficiently due to limited data availability in this field.
This study, due to data accessibility, focuses on the analysis of upper limb bradykinesia among 5 actions, Finger Tapping, Hand Movement, and Rapid Alternating Movements (AM), observing that existing research does not clearly represent features indicative of "occasional arrest"(arrest) and "gradual decrement in amplitude"(fatigue) when performing repeated movements[5],[11]-[13]. Previous studies have implicitly included features like interval range, average jerk, rate of inversion[13], or mean and variance of the interval to account for arrests[5], which may not accurately reflect the criteria of MDS-UPDRS for arrest and fatigue. Therefore, this research employs an LSTM-based architecture, LSTM-FCN[16], suitable for time series analysis, to classify motion for quantification of arrest. Moreover, a decrement in amplitudes is captured by numerical analysis, creating fatigue features. As shown in Figure 1, These are then used in conjunction with other features derived from numerical analysis in a hierarchical classification structure. This approach has been applied to a dataset of 1396 videos, demonstrating robust results across diverse data. Effective significance test by semiparametric additive model[17] has been implemented to test whether created features are significant in analyzing bradykinesia severity.

\newpage

\section{Related Works}

Previous studies on the quantification of bradykinesia using computer vision have commonly employed human pose estimation to generate motion graphs, from which features are extracted and then processed through an ordinal classifier to produce scores.

\bigskip

\textbf{Human Pose Estimation}

\smallskip

While earlier research focused on lightweight alternatives to OpenPose's heavier model[14], the field has evolved with the advent of state-of-the-art technologies like Google's MediaPipe[18]. MediaPipe offers the dual advantages of convenience and accuracy, even on mobile platforms. In this study, Google MediaPipe was employed for human pose estimation.

\bigskip

\textbf{Feature Extraction and Ordinal Classification}

\smallskip

Liu et al.[5] extracted features such as amplitude and interval means and variances from smoothed data by identifying local extrema. However, these features alone do not sufficiently capture essential aspects of bradykinesia, such as Early Fatigue (Decrease in Amplitude) and Occasional Arrest (Arrest in Movement), making it challenging to derive robust features for diverse patient data. Morinan et al.[13] attempted an improvement by selecting a total of 11 features, which included indications for arrest and decremental signals. However, their focus on arrest emphasized interval volatility rather than direct hesitancy, and the decremental signal lacked clarity regarding the specific timing of occurrence. Consequently, these features did not explicitly describe the assessment criteria, leading to potential misjudgments.

\smallskip

While there have been attempts to employ deep learning for the quantification of PD symptoms[19]-[20], a question still remains regarding the optimal approach for quantification of bradykinesia. Given the precise and well-established criteria of the MDS-UPDRS , some researchers argue in favor of a hybrid model approach[13]. This perspective suggests integrating both deep and shallow machine learning models, rather than relying solely on fully deep learning architectures. The rationale behind this is that combining the complex deep learning model with the simple and interpretable shallower model could more effectively meet the specific requirements and nuances of the MDS-UPDRS criteria.

\smallskip

In addressing these challenges, some studies[21] utilized DNN with unsupervised classification, but clustering approaches can be ineffective in dealing with severity classification. Our study introduces refined indicators via local regression to accurately measure Early Fatigue, a key aspect of bradykinesia. To capture instances of arrest, we deployed an LSTM-FCN model, which excels in classifying complex time-series data[16]. This quantified arrest data was then incorporated into a comprehensive hierarchical classification framework. This framework not only leverages the depth of LSTM-FCN but also integrates it with traditional ordinal classification methods, creating a nuanced and multifaceted approach to bradykinesia assessment. 

\smallskip

Furthermore, unlike other studies[5], [13] that relied on basic correlation related analysis to justify the selected features, significance of arrest/fatigue related features are tested by semiparametric additive regression model, which shows effective performance in estimation[17]. By this testing, our study demonstrated not only the importance of precise quantification of fatigue and arrest, but also the performance of our proposed features.

\bigskip

\bigskip

\section{Methods}
\begin{figure*}[t]
\includegraphics[width=\textwidth]{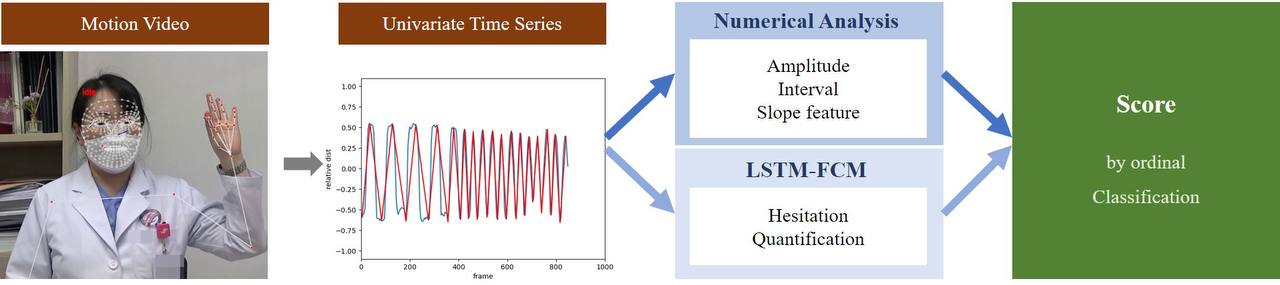}
\caption{\emph{This figure illustrates the comprehensive pipeline of our approach. Utilizing Google MediaPipe[18], a state-of-the-art technology for human pose estimation, we extract a graph from the motion video. The video is then analyzed to create numerical features. Concurrently, the LSTM-FCN model is trained to quantify arrest. Based on the outputs of this stage, we perform ordinal classification to construct a score sheet.}}
\end{figure*}
This study focuses on the quantification of bradykinesia in PD using three hand-related movement scales
of MDS-UPDRS 3.5 to 3.7: Finger Tapping, Hand Movement, and Rapid AM. Videos of patients performing these actions serve as the primary input.

\vspace{10mm}

\textbf{Hand Pose Estimation}

\smallskip

Unlike some studies which develop their own networks[5], we utilized Google MediaPipe for human pose estimation. The holistic model of MediaPipe was employed for more stable hand joint detection. The extracted image coordinates were rescaled into pixel coordinates to represent the actual distances between hand feature points, normalized by each patient's palm length for comparative purposes. Some studies have estimated the height of patients to use as a normalizing factor[23]-[24]. However, in many cases, only the upper body is captured in the recordings, and since the focus is on the relative movements of the hands, we chose to normalize using the length of the palm. This approach allows for a more accurate assessment of hand movements by accounting for individual variations in hand size, ensuring that the analysis is based on proportional movement scales. For the calculation of palm length, owing to stability issues, we used the 'thumb\_cmc' point instead of the wrist point. Furthermore, for Rapid AM, distances were converted to negative values when palms faced the body, for the sake of accurate representation of the movement. Extracted features can be found in Table 2.

\bigskip

\begin{center}
\includegraphics[scale=0.4]{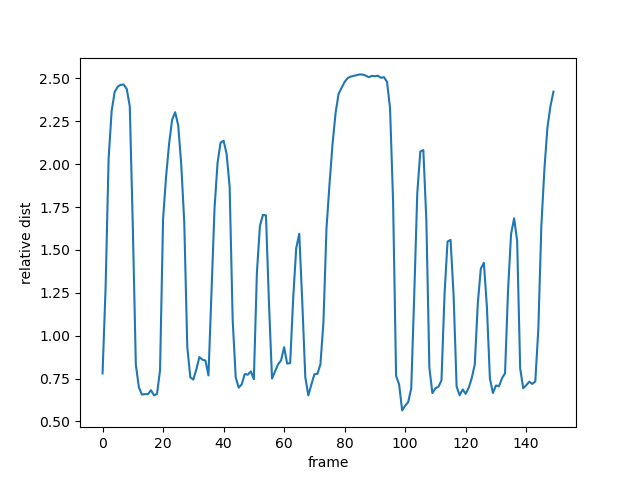}\includegraphics[scale=0.4]{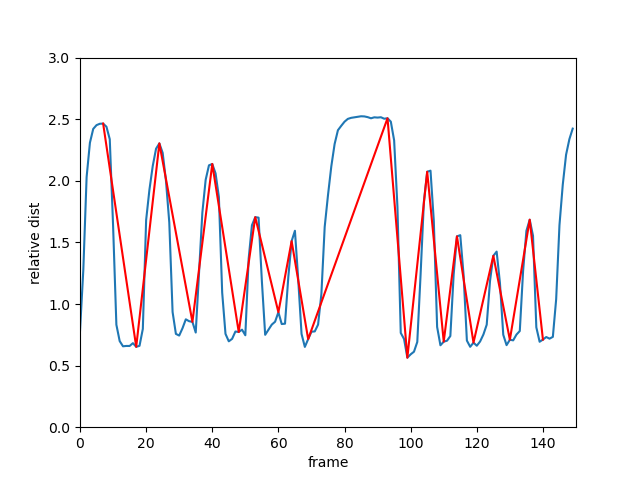}
\end{center}~~\\
Figure 2. \emph{Example of Hand Movement time series graph. Among many graphs with high volatility in amplitude and interval, this graph shows early fatigue and severe arrest, which cannot be captured by previous studies. The red line illustrates the local extrema. We employed the Savitzky-Golay filter for this purpose, with hyperparameters set variably according to the type of movement. Specifically, the window size and polynomial order were configured as (7,3) for Finger Tapping, (7,3) for Hand Movements, and (5,4) for Rapid AM. A thorough examination of the total dataset was conducted to eliminate any false extrema.}

\vspace{18mm}

\textbf{Graph Feature Extraction}

\bigskip

The pose estimation data were transformed into graphs and then processed for feature extraction. Six features were constructed based on amplitude and interval, aligned with MDS-UPDRS criteria. Inspired by previous researches[5], [25], we applied a Savgol filter for identifying local extrema and forming amplitude and interval values. The first 10 movements were specifically analyzed, considering the MDS-UPDRS's[2] focus on initial actions. False extrema were filtered out after thorough dataset verification. Example of graph preprocessing can be found in Figure 2.

\begin{table*}
\includegraphics[scale=0.6]{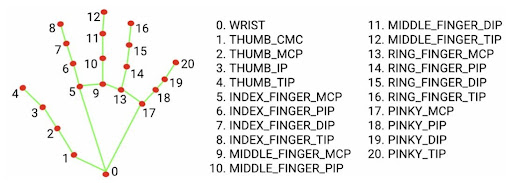}\\
~~\\
\begin{tabular}{clc}
\hline
Feature Type & description & Fromulation \\
\hline
Palm Length & \begin{tabular}{l}
length of palm
\end{tabular} 
& $|x_{1}-x_{9}|$\\
\hline
Finger Tappling & \begin{tabular}{l}
Euclidean distance between tips of thumb \\
and index finger 
\end{tabular} & $\cfrac{|x_{4}-x_{8}|}{|x_{1}-x_{9}|}$\\
\hline
Hand Movement & \begin{tabular}{l}
Euclidean distance between middle finger \\
tip and wrist 
\end{tabular} & $\cfrac{|x_{1}-x_{12}|}{|x_{1}-x_{9}|}$\\
\hline
Rapid AM & \begin{tabular}{l}
Difference between $2d$ coordinates of 
index and little finger 
\end{tabular}  & $\cfrac{|x_{5}[:2]-x_{17}[:2]|}{|x_{1}-x_{9}|}$\\
\hline
\end{tabular}\\
~~~\\
 {\raggedright $x_{i}$=coordinate of the $i$-th point which can be found above\par}.
    \caption{\emph{The figure above showcases the feature set provided by MediaPipe[17], consisting of a 20-point skeleton set. The table depicts the features extracted for three distinct movements. The corresponding points are extracted frame by frame, thereby constructing a univariate time series for each movement.}}
\end{table*}

\bigskip

\textbf{Fatigue Feature (Early Fatigue)}

\smallskip

The fatigue feature was designed to quantify the aspect of the bradykinesia criteria that assigns higher scores when fatigue, characterized by a decrease in amplitude, occurs more rapidly. Zhao et al.[21] implemented clustering approach to tackle this task, but our study tries to quantify severity by numerical approach. In this study, local regression with a window size of 5 was used for amplitude data to represent the decrement in amplitude. A significance test of significance 0.1 on the linear regression coefficient was conducted, factoring in the window index to enhance the feature's impact on scoring.

\begin{displaymath}
\mathrm{fatigue}\_\mathrm{feature}(x)=\sum_{i}-\frac{\beta^{3}}{\sqrt{i+1} \big(\frac{\overline{\mathrm{loc\_amp}} }{\overline{\mathrm{amp}}}\big)^{2} }   
\end{displaymath}
$x$ : amplitude vector \\
$\beta$ : coefficient of local regression \\
$i$ : window\_index \\
loc\_amp = average value of local amplitude\\
amp = average value of amplitude for total motion

\newpage

\textbf{Hierarchical Classification}

\smallskip

The arrest feature classification was executed using an LSTM-FCN model, chosen for its proven effectiveness in handling time series data and its capability to discern occasional arrests in movements[26]. To adapt to the small size of our dataset, we normalized amplitude and interval values for each action before feeding them into the network. This process was supplemented by supervised learning, guided by insights from medical experts, resulting in the development of a refined scoring chart.

\par

In our model, the LSTM-based function $f$ maps normalized time series graphs to an arrest score, formulated as:
\begin{align*}
    f &: \mathcal{G} \to \{0,1,2,3\} \\
\end{align*}
Here, $\mathcal{G}$ denotes the set of normalized time series graphs, with the range of arrest scores confined to $\{0, 1, 2, 3\}$.

\par

Subsequently, for the final bradykinesia score, we utilized the XGBoost classifier[27]. XGBoost was selected due to its superior performance in gradient boosting methods and its compatibility with the OR-gate logic of the MDS-UPDRS bradykinesia criteria, as detailed in Table 1.

\par

The function $g$, implemented using XGBoost, employs a feature set comprising mean and relative standard deviation (rsd) of amplitude and interval, along with fatigue and arrest scores, to determine the final bradykinesia score:
\begin{align*}
    g &: \mathcal{F} \to \{0,1,2,3\} \\
\end{align*}
In this context, $\mathcal{F}$ symbolizes the feature space utilized for bradykinesia scoring.

\bigskip

\textbf{Statistical Analysis by Semiparametric Additive Model}

\smallskip

To check the need of arrest/fatigue related features and performance of our proposed features, careful statistical analysis has been implemented. Among many analysis schemes, partially linear additive regression has been used based on its efficacy in coefficient estimation[17]. Unlike typical regression problems, the response variable in this study is ordered categorical value, therefore the link function works to model the cumulative probability that the response falls in or below each category.\\
\vspace{-0.3cm}
\begin{align*}
    g(\mu) &= \beta_0 + s_1(\text{mean\_amp}) + s_2(\text{rsd\_amp})+ s_3(\text{mean\_int}) + s_4(\text{rsd\_int}) \\
    &\quad + \beta_1 \times \text{fatigue\_feature} + \sum_{k=0}^{3} \gamma_k \times I(\text{arrest\_feature} = k)
\end{align*}

\textit{where:
\begingroup
\setlength{\itemsep}{0pt}
\setlength{\parskip}{0pt}
\begin{itemize}
    \item $g(\cdot)$ is the cumulative logit link for ordinal responses.
    \item $\mu$ is the expected value of the ordinal response variable.
    \item $\beta_0$ is the fixed intercept.
    \item $s_k(\cdot)$ are smooth functions for the $k$-th predictor. The term amp and int are representing amplitude and interval, respectively.
    \item $\beta_1$ is the coefficient for the linear predictor \text{fatigue\_feature}.
    \item $\gamma_k$ are coefficients for the factor variable \text{arrest\_feature}.
\end{itemize}
\endgroup
}

During the estimation process, continuous smooth backfitting is applied to maximize the smoothness–penalized likelihood until the convergence. Once parameters are converged, coefficients are extracted and similar processes by bootstrapping can be done to estimate the standard error of coefficients. Compared to original linear model and other semiparametric models, this model is efficient and one-dimensional rate of convergence is assured[17]. By this process, p-values for the significance testing of coefficients of parametric part can be calculated.

\vspace{0.3cm}

\section{Experiments}

\textbf{Patients and Data Preparation}

\smallskip

This study incorporated videos from a total 310 patients diagnosed with PD at the Movement Clinic of Hallym University Sacred Heart Hospital. The recording captured motor features based on the MDS-UPDRS part III.
Our dataset comprises 2015 videos, taken by 2D camera, including 1396 videos and 619 records from healthy individuals. The videos from healthy individuals are utilized only for training of the LSTM-FCN model. For a comprehensive framework, we excluded the data with score 4 due to its limited sample size and often impractical filming conditions[22].

\smallskip

The videos were recorded with informed consent at the time of filming as part of the medical record and academic purpose. Additional informed consent specifically for video analysis was waved due to the retrospective nature of the study and the absence of privacy exposure risk. The Institutional Review of Board (IRB) of Hallym University Sacred Heart Hospital approved this study (IRB no.\textbf{2023-12-010}) for ethical considerations.

\smallskip

Three hand-related motor movements underwent review and consensus by three movement specialists(HM, YK, and GH) according to the UK Brain Bank Criteria[28] and the score dataset can be found in Table 3. Given that the dataset consists primarily of patient data, there is a relatively low prevalence of normal scores and a higher prevalence of severe scores. Thus, inspired by G. Morinan et al.[13], we employed stratified split testing and used the Area under the ROC curve as our primary metric.

\smallskip

\begin{center}
\begin{tabular}{cccccccc|}
\hline
score  & \begin{tabular}{l}
Finger \\
Tapping \\
\end{tabular} & \begin{tabular}{l}
Hand \\
Movement \\
\end{tabular} & \begin{tabular}{l}
Rapid\\
AM \\
\end{tabular} \\
\hline
0 & $\mathrm{L}: 10$ & $\mathrm{~L}: 26$ & $\mathrm{~L}: 22$ \\
 & $\mathrm{R}: 6$ & $\mathrm{R}: 23$ & $\mathrm{R}: 33$ \\
\hline
1 & $\mathrm{L}: 59$ & $\mathrm{~L}: 49$ & $\mathrm{~L}: 87$ \\
 & $\mathrm{R}: 63$ & $\mathrm{R}: 61$ & $\mathrm{R}: 74$ \\
\hline
2 & $\mathrm{L}: 106$ & $\mathrm{~L}: 76$ & $\mathrm{~L}: 57$ \\
 & $\mathrm{R}: 122$ & $\mathrm{R}: 77$ & $\mathrm{R}: 82$ \\
\hline
3 & $\mathrm{L}: 55$ & $\mathrm{~L}: 71$ & $\mathrm{~L}: 59$ \\
  & $\mathrm{R}: 50$ & $\mathrm{R}: 72$ & $\mathrm{R}: 47$ &  \\
\hline
\end{tabular}
\hspace{0.3cm}
\begin{tabular}{|ccc|} 
\hline
\multicolumn{3}{|c|}{\textbf{Patient Demography}} \\
\hline
\toprule
{Category} & {Male (M)} & {Female (F)} \\
\hline
\midrule
Count       & 147           & 164          \\
Mean Age    & 65.6         & 67.7        \\
Med Age  & 67.0         & 69.0        \\
SD Age & 11.0         & 9.92        \\
Min Age     & 35           &  37         \\
Max Age     & 89           & 86          \\
\bottomrule
\hline
\end{tabular}
\end{center}
Table 3. \emph{This table displays patient demography and the score distribution of the patient videos utilized in the experiment. A total of 1396 videos were used, collected from 310 patients who participated in the data gathering process.}

\bigskip

\textbf{Results}

\smallskip

The results are visualized sequentially, considering the independent learning processes of the LSTM-FCN and the higher-level classifier. Confusion matrices and ROC curves were generated for each action.

\bigskip

\emph{LSTM-FCN for arrest feature}

\smallskip

The LSTM-FCN model was trained using a comprehensive dataset that encompassed both patients with PD and healthy individuals. This inclusive approach ensured that the model learned to differentiate various movement patterns effectively, critical for accurate bradykinesia assessment.\\
In total, 2015 videos were incorporated into the training process. The training process aimed to finely tune the LSTM-FCN model so that it could accurately identify and quantify instances of arrest or arrest in movements, characteristics symptomatic of bradykinesia in PD[2]. LSTM units in the model are adept at processing time-series data, capturing the temporal dynamics of the hand motions, while the FCN part effectively manages the spatial aspects of the motion captured in the videos.

\newpage

\begin{multicols}{2}

\centering
LSTM - FCN ${}_{{}_{}}$
\includegraphics[scale=0.4]{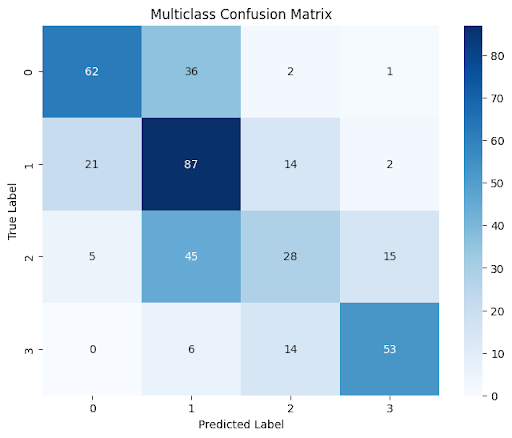}
\captionof*{figure}{Figure 3-a. \emph{Confusion Matrix for the result of training LSTM-FCN in arrest quantification}}

\vspace{10mm}

\begin{center}
\includegraphics[scale=0.4]{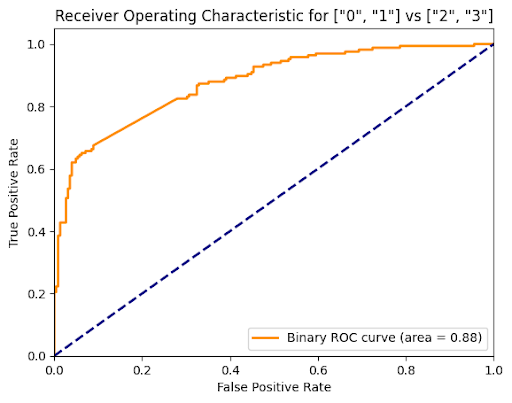}
\captionof*{figure}{Figure 3-b. \emph{ROC curve for binary classification of mild and severe arrest.}}
\end{center}

\footnote{Figure 3-a and 3-b. Analysis of LSTM-FCN Results with Confusion Matrix and ROC Curve\\
The results of the LSTM-FCN model, as depicted in the Confusion Matrix, primarily cluster around the central axis, indicating a high degree of accuracy in predictions. When considering a margin of error of $\pm$ 1 is acceptable even between experts[29], the fitting appears to be quite precise, signifying that the model is generally accurate in classifying the correct scores for the patient videos. Most entries in the Confusion Matrix are concentrated around the diagonal line, which represents accurate predictions where the model’s estimated scores align with the actual scores. This distribution suggests that the model is not only accurate but also consistent in its performance across different levels of bradykinesia severity.
Furthermore, the Area Under the ROC (Receiver Operating Characteristic) Curve for the model is 0.88. This metric is significant as it encapsulates the model's ability to discriminate between different classes effectively. An AUC[30] (Area Under Curve) of 0.88 is indicating that the LSTM-FCN model has a high probability of correctly distinguishing between varying severities of bradykinesia, as well as between movements of PD patients and those of healthy individuals.}

\begin{center}
Finger Tapping ${}^{}$
\includegraphics[scale=0.4]{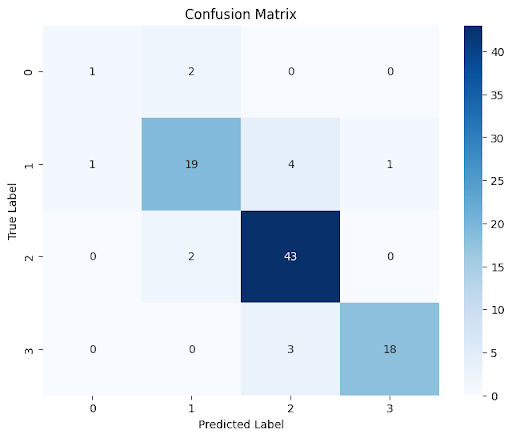}
\captionof*{figure}{Figure 4-a. \emph{Confusion Matrix for the score of Finger Tapping}}
\end{center}

\vspace{14mm}

\begin{center}
\includegraphics[scale=0.4]{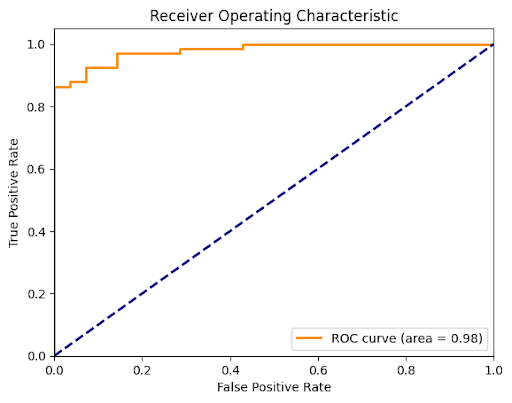}
\captionof*{figure}{Figure 4-b. \emph{ROC curve for binary classification of Finger Tapping related bradykinesia severity.}}
\end{center}

\newpage

\begin{center}
Hand Movement ${}_{{}_{}}$
\includegraphics[scale=0.4]{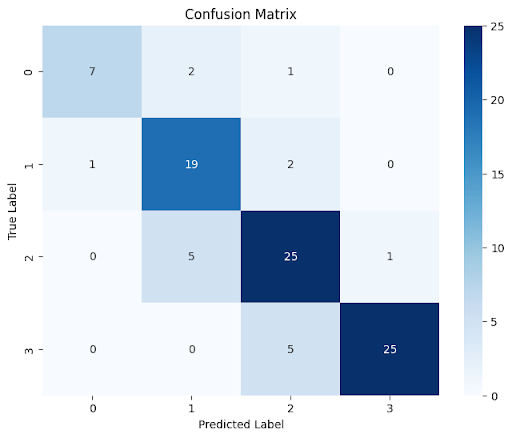}
\captionof*{figure}{Figure 5-a. \emph{Confusion Matrix for the score of Hand Movement}}
\end{center}

\vspace{10mm}

\begin{center}
\includegraphics[scale=0.4]{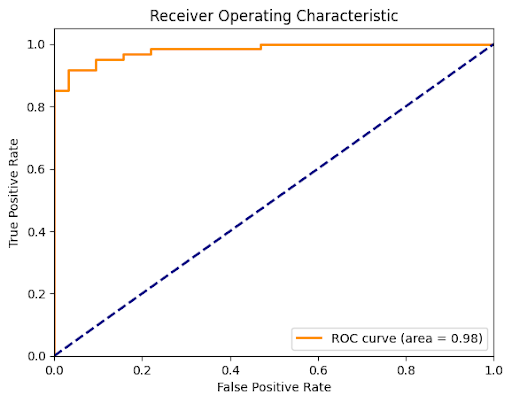}
\captionof*{figure}{Figure 5-b. \emph{ROC curve for binary classification of Hand Movement related bradykinesia severity.}}
\end{center}
\footnote{Figure 4-a,b, 5-a,b, 6-a,b. Performance Analysis for the Three Types of Movements\\
Following the LSTM-FCN results, the analysis for each of the three specific movements - Finger Tapping, Hand Movement, and Rapid AM - showed remarkably high overall accuracies. The overall accuracy for Finger Tapping was found to be 86.2\%, for Hand Movement it was 81.7\%, and for Rapid AM, the accuracy was 75.7\%. These figures indicate a high level of precision in the model's ability to correctly classify the movements, which is crucial for the reliable assessment of bradykinesia in PD.\\
Moreover, all three movements demonstrated an Area Under the ROC (Receiver Operating Characteristic) Curve of 0.98. ROC curve is for binary classification between mild and severe ([0,1] vs[2,3]). Performance on the binary task was comparable to previous studies[12]-[13], [31]. This exceptionally high AUC value suggests that the model has an excellent capability to differentiate between different severity levels of movements accurately. An AUC of 0.98 implies that the model has a very high true positive rate while maintaining a low false positive rate, showcasing its effectiveness in distinguishing between the nuanced movements typical of PD patients and those of healthy individuals.
}
\columnbreak 

\begin{center}
Rapid AM ${}_{{}_{}}$
\includegraphics[scale=0.45]{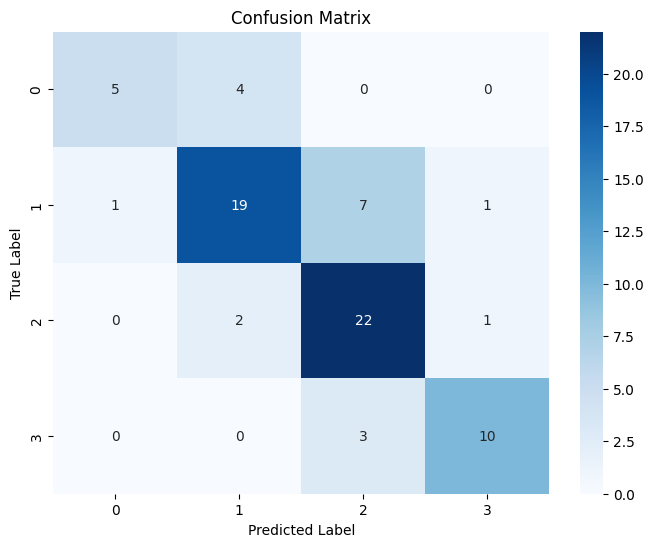}
\captionof*{figure}{Figure 6-a. \emph{Confusion Matrix for the score of Rapid AM}}
\end{center}

\vspace{10mm}

\begin{center}
\includegraphics[scale=0.48]{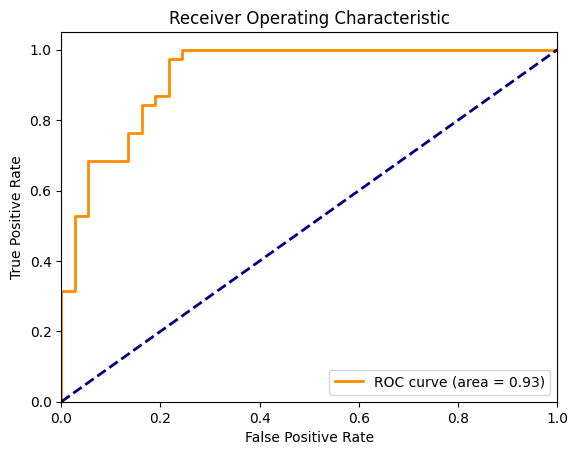}
\captionof*{figure}{Figure 6-b. \emph{ROC curve for binary classification of Rapid AM related bradykinesia severity.}}
\end{center}
\end{multicols}

\begin{table*}[ht]
\centering
\label{tab:my_label}
\begin{tabular}{@{}lccccc@{}}
\toprule\hline
        & Fatigue & Arrest cat 1 & Arrest cat 2 & Arrest cat 3 & Deviance explained \\ \midrule\hline
FT Left  & $<0.001$                 & 0.017                          & $<0.001$                       & 0.137                          & 69\%                         \\
\hline
FT Right & $<0.001$                 & 0.864                          & 0.486                          & 0.335                          & 91.9\%                       \\
\hline
HM Left  & 0.906                    & 0.268                          & 0.465                          & $<0.001$                       & 84.3\%                       \\
\hline
HM Right & 0.506                    & 0.032                          & 0.495                          & 0.558                          & 75.7\%                       \\
\hline
RA Left  & $<0.001$                 & 0.098                          & 0.193                          & 0.118                          & 92.7\%                       \\
\hline
RA Right & 0.03                     & 0.251                          & 0.877                          & 0.752                          & 88.4\%                       \\ 
\hline
\bottomrule
\end{tabular}\\
\vspace{0.2cm}
\raggedright
Table 4. \emph{This table shows the p value result of coefficient significance test of fatigue and arrest feature for partially linear additive model. For each action, because of data dependency, extracted significant features are different.}
\end{table*}

\newpage

\emph{Hierarchical Classifier}

\smallskip

The feature maps and scores were divided using a 5-fold split. For each feature map, the arrest feature was assigned by score from clinical experts to test the performance of each classifier. Based on this configuration of features and scores, Confusion Matrices and ROC Curves were constructed for each action as shown in Figure 6 to Figure 8.

\bigskip

\emph{Ablation Study}

\bigskip

\begin{center}
\begin{tabular}{ccc}
\hline
 & \begin{tabular}{l}
Overall Acc \\
(full/ablated ) \\
\end{tabular} & \begin{tabular}{l}
Area under \\
ROC Curve \\
(full/ablated) \\
\end{tabular} \\
\hline
\begin{tabular}{l}
Finger Tapping
\end{tabular} & \begin{tabular}{l}
$86.2 \% / 79.8\%$ \\
\end{tabular} & $0.98 / 0.87$ \\
\hline
\begin{tabular}{l}
Hand Movement
\end{tabular}  & \begin{tabular}{l}
$81.7 \% / 77.4 \%$ \\
\end{tabular} & $0.98 / 0.89$ \\
\hline
\begin{tabular}{l}
Rapid AM
\end{tabular}  & \begin{tabular}{l}
$75.7 \% / 67.9\%$ \\
\end{tabular} & $0.93 / 0.81$ \\
\hline
\end{tabular}
\end{center}

Table 5. \emph{Ablation Study Outcomes for Bradykinesia Movement Classification}

\bigskip

The result from the ablation study shown in Table 5 reveal the significance of the arrest/fatigue features in the model's performance. Ablated model lacks arrest/fatigue features proposed by our study. Particularly for Finger Tapping and Rapid AM, the more considerable decrease in both accuracy and AUC in the ablated model underscores the importance of these features in accurately classifying the movements associated with PD. The result illustrates the importance of arrest/fatigue features in bradykinesia quantification, which can be further solidified by statistical analysis below.

\bigskip

\textbf{Statistical Analysis}

\smallskip

\emph{Partially Linear Additive Model for feature significance test}

\smallskip

As shown in Table 4, the statistical significance of the fatigue feature is evident for Finger Tapping and Rapid AM, but it does not emerge as a significant factor for Hand Movement. This discrepancy underscores that the key features for quantifying bradykinesia can vary depending on the specific action being assessed.

\smallskip

While the significance of arrest features can vary
due to data dependency and the logic employed
by the MDS-UPDRS criteria, which employs an
OR-gate driven structure, it becomes evident that
certain features can exert dominance depending on
the context, resulting in differing levels of statistical significance. However, even when considering these variations, our proposed fatigue and arrest features often exhibit statistically significant associations in many instances. This suggests that, in the context of bradykinesia quantification, fatigue and arrest features can be deemed statistically significant factors, given their frequent significant presence.

\newpage

\textbf{Discussion}

\bigskip

\emph{Mixed Effects Model for Score Comparison}

\smallskip

\begin{align*}
    Y_{ij} = \beta_0 + u_{0j} + \epsilon_{ij}
\end{align*}

\textit{Where:}
\begingroup
\setlength{\itemsep}{0pt}
\setlength{\parskip}{0pt}
\begin{itemize}
    \item $Y_{ij}$ \textit{represents the score for the $i$-th observation in the $j$-th task group. This score is the difference between the predicted score and the expert score for that particular observation and task.}
    \item $\beta_0$ \textit{is the fixed intercept, indicating the overall average score difference across all task groups.}
    \item $u_{0j}$ \textit{is the random effect associated with the $j$-th task group, which accounts for the variation in score differences across different tasks:}
    \begin{itemize}
        \item $u_{01}$ \textit{for the "Finger Tapping" task group.}
        \item $u_{02}$ \textit{for the "Hand Movement" task group.}
        \item $u_{03}$ \textit{for the "Rapid AM" task group.}
    \end{itemize}
    \textit{\item $\epsilon_{ij}$ is the random error for the $i$-th observation in the $j$-th task group.}
\end{itemize}
\endgroup
In this study, we treated the bradykinesia scores assigned by medical experts as samples extracted from a 'true score set'. The primary objective was to develop a diagnostic machine capable of substituting medical experts' evaluations, especially when considering the practical applications of the research. Consequently, we based our analysis on these expert scores, deriving a confusion matrix and AUC-ROC for assessment.

\smallskip

However, from a statistical perspective, it is imperative to validate the significant difference between two ordered categorical samples. For this purpose, our research employed a mixed-effects model, defined by the formula above, to verify the significant differences between scores given by experts and those determined by our diagnostic machine.

\smallskip

Fitting the model above, we yield 0.557 for p-value of significance of $\beta_0$, therefore there isn't a statistically significant difference between prediction and score in practice.

\bigskip

\emph{Further Research}

\smallskip

The outcomes of this study, particularly the high accuracy and area under the ROC curve for each of the tested movements, are indeed impressive compared to previous studies[12]-[13]. Such results demonstrate the potential effectiveness of our model in accurately classifying and quantifying bradykinesia in PD. The LSTM-FCN model, supplemented with robust feature sets, shows considerable promise in distinguishing between the nuanced movements characteristic of PD and those of healthy individuals. Considering other network models like TCN[32], transformer-based algorithms can be effective in building precise feature, which remains to be conducted in our future research.

\smallskip

It's also important to acknowledge that the majority of our dataset comprised videos recorded in a single hospital. While this provided a controlled environment for data collection and initial model training, it also presents a limitation in terms of the diversity[13] and generalizability of our findings. A dataset predominantly sourced from one location may not fully represent the variability in PD symptoms and movements seen across broader populations.

\smallskip

Despite these considerations, the high performance of the model, particularly in terms of accuracy and the area under the ROC curve, is noteworthy. Such results suggest that the model is highly effective in capturing the essential characteristics of bradykinetic movements, a critical aspect of PD assessment. This claim can be supported by the result of significance testing of the partially linear additive model’s parametric part, which indicates that our proposed features are effective in bradykinesia severity modeling.

\newpage

\section{Conclusion}
In this study, we introduced hierarchical classification based on proposed arrest/fatigue features to accurately model bradykinesia severity in PD. The arrest feature is created by deep learning, and the fatigue feature is drawn by careful numerical analysis. These features, proven statistically significant through semiparametric additive modeling, showed effective performance in the classification scheme.  This approach demonstrated robust effectiveness in bradykinesia quantification, marking a significant advancement in the precise assessment of this condition.

\bigskip
\bigskip

\textbf{Code and Data Availability}

\bigskip

This study, a collaborative effort between VTOUCH and Hallym Sacred Heart Hospital, was funded by the Korean government. Code and data availability can be inquired from the authors under appropriate circumstances.

\end{document}